\documentclass[3p,12pt]{elsarticle}

\usepackage{lineno}
\usepackage{amsmath}
\usepackage{url}

\usepackage{times}
\usepackage{epsfig}
\usepackage{graphicx}
\usepackage{amsmath}
\usepackage{amssymb}
\usepackage{color}

\usepackage{algorithmicx}
\usepackage{algorithm}
\usepackage{algpseudocode}
\usepackage{multirow, makecell}
\usepackage{float}

\journal{Journal of Computational Science}

\begin{document}
	
	\begin{frontmatter}
		
		%% Title, authors and addresses
		
		%% use the tnoteref command within \title for footnotes;
		%% use the tnotetext command for theassociated footnote;
		%% use the fnref command within \author or \address for footnotes;
		%% use the fntext command for theassociated footnote;
		%% use the corref command within \author for corresponding author footnotes;
		%% use the cortext command for theassociated footnote;
		%% use the ead command for the email address,
		%% and the form \ead[url] for the home page:
		%% \title{Title\tnoteref{label1}}
		%% \tnotetext[label1]{}
		%% \author{Name\corref{cor1}\fnref{label2}}
		%% \ead{email address}
		%% \ead[url]{home page}
		%% \fntext[label2]{}
		%% \cortext[cor1]{}
		%% \affiliation{organization={},
		%%             addressline={},
		%%             city={},
		%%             postcode={},
		%%             state={},
		%%             country={}}
		%% \fntext[label3]{}
		
		\title{
		Boundary Learning by Using Weighted Propagation in Convolution Network
		% Deep learning based material microscopic image segmentation combing 3D information
		}
        % WPU-Net: Boundary Learning by Using Weighted Propagation in Convolution Network
		\author[BAICMGE,SCCE,SGSUSTB,BKLKE,CUGTT]{Wei Liu \fnref{EC}}
		\author[BAICMGE,SCCE,SGSUSTB,BKLKE]{Jiahao Chen \fnref{EC}}
		\author[SCCE,PCG]{Chuni Liu \fnref{EC}}
		\author[BAICMGE,IAT,IAMT,SGSUSTB,BKLKE]{Xiaojuan Ban \corref{BCA}}
		\author[BAICMGE,IAT,IAMT,SGSUSTB,BKLKE]{Boyuan Ma \corref{MBY}}
		\author[SMSE]{Hao Wang}
		\author[SMST,BKLKE]{Weihua Xue}
		\author[BAICMGE,SCCE,SGSUSTB,BKLKE]{Yu Guo}
		
		\address[BAICMGE]{Beijing Advanced Innovation Center for Materials Genome Engineering, University of Science and Technology Beijing, China.}
		\address[IAT]{Institute of Artificial Intelligence, University of Science and Technology Beijing, China.}
		\address[SCCE]{School of Computer and Communication Engineering, University of Science and Technology Beijing, China.}
		\address[IAMT]{Institute for Advanced Materials and Technology, University of Science and Technology Beijing, China.}
		\address[SGSUSTB]{Shunde Graduate School of University of Science and Technology Beijing, China.}
		\address[PCG]{Platform and Content Group, Tencent, China.}
		\address[SMSE]{School of Materials Science and Engineering, University of Science and Technology Beijing, China.}
		\address[SMST]{School of Materials Science and Technology, Liaoning Technical University, China.}
		\address[BKLKE]{Beijing Key Laboratory of Knowledge Engineering for Materials Science, China.}
		\address[CUGTT]{China United Gas Turbine Technology CO.,LTD, China.}

		\fntext[EC]{These authors contributed equally to the work.}
		\cortext[BCA]{corresponding authors: banxj@ustb.edu.cn.}
		\cortext[MBY]{corresponding authors: mbytony@ustb.edu.cn.}
		
		\begin{abstract}
			%% Text of abstract
			In material science, image segmentation is of great significance for quantitative analysis of microstructures. Here, we propose a novel Weighted Propagation Convolution Neural Network based on U-Net (WPU-Net) to detect boundary in poly-crystalline microscopic images. We introduce spatial consistency into network to eliminate the defects in raw microscopic image. And we customize adaptive boundary weight for each pixel in each grain, so that it leads the network to preserve grain's geometric and topological characteristics. Moreover, we provide our dataset with the goal of advancing the development of image processing in materials science. Experiments demonstrate that the proposed method achieves promising performance in both of objective and subjective assessment. In boundary detection task, it reduces the error rate by 7\%, which outperforms state-of-the-art methods by a large margin.
		\end{abstract}
		
		%%Research highlights
		%%\begin{highlights}
		%%	\item An end-to-end network can simultaneously generate decision map and fused result
			
		%%	\item A loss function can preserve gradient information and improve fusion quality
			
		%%	\item A decision calibration strategy can increase efficiency for multiple images fusion
		%%\end{highlights}
		
		
	\end{frontmatter}
	
% 	\linenumbers
	
	%% main text
	\section{Introduction}
	\label{sec:introduction}
	Material microstructures are determined by material composition and preparation process, which is of great significance for controlling the properties and performance of materials~\cite{hu2017grain, ban2020, ma2021deep}. And most metals and ceramics are composed of complex microstructures, which separated by different interfaces such as grain boundary~\cite{forsyth1946grain}, phase boundary~\cite{jagitsch1947a} and domain boundary~\cite{chou1982anti-phase}. Therefore, boundary detection in microscopic image is one key step to quantitatively analyze the material structure and estimate the material properties.

    \begin{figure}[H]
    \centering
    \includegraphics[width=0.8\linewidth]{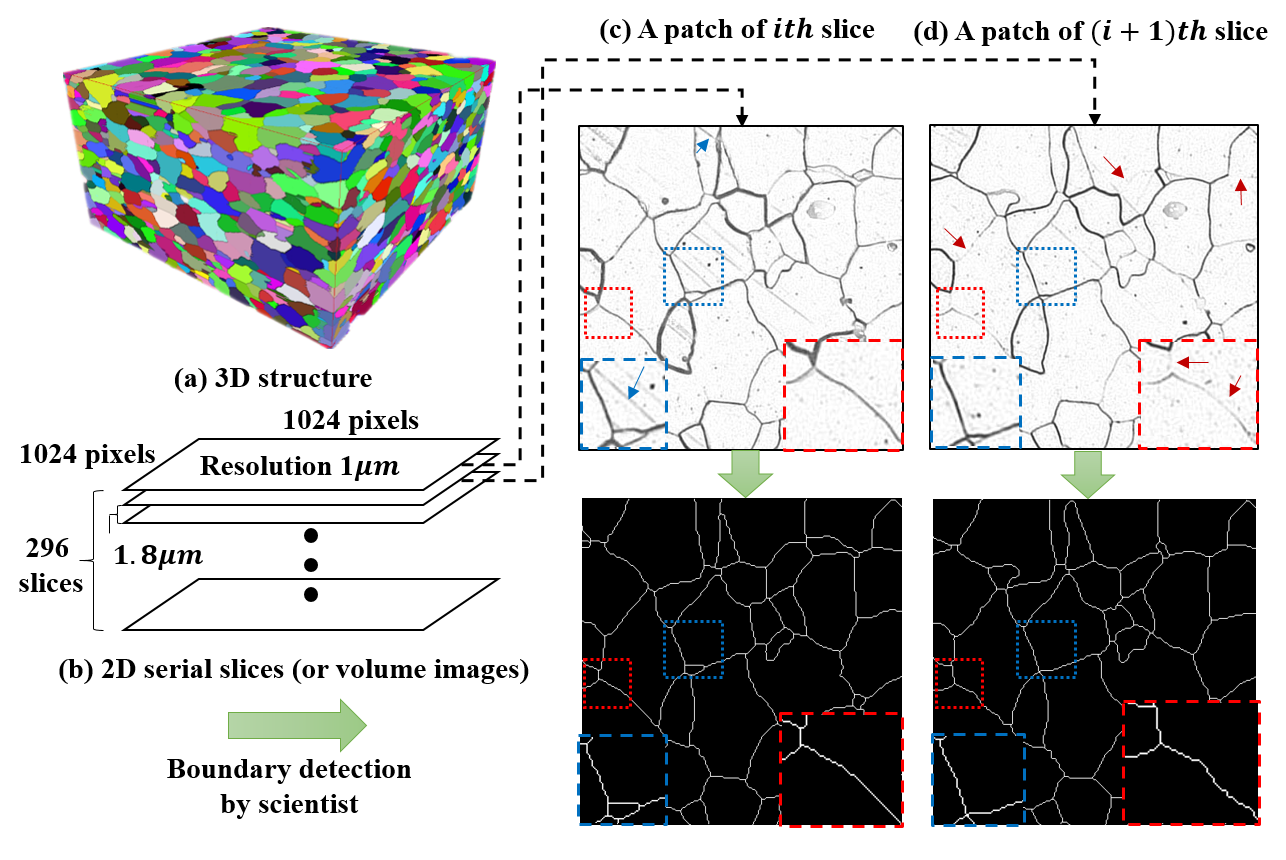}
    \caption{Microscopic images of polycrystalline iron. (a) The 3D structure; (b) 2D serial slices (or volume images); (c) A patch of $i$th slice; (d) A patch of $(i+1)$th slice. There are some defects appeared in raw image, such as blurred or missing boundary (red arrows and one region is magnified and shown in the lower right corner with red rectangle) and scratches (blue arrows and one region is magnified and shown in the lower left corner with blue rectangle).}
    \label{microscopic_images}
    \end{figure}
    
    Taking poly-crystalline iron for example, the ultimate objective is to obtain and characterize the 3D material structure, shown in Figure \ref{microscopic_images}(a). Due to the opacity of that, researchers can only use serial section method to obtain 2D serial slices (or volume images), shown in Figure \ref{microscopic_images}(b), and stack these to reconstruct 3D structure. Among them, 2D segmentation is used as the pre-step of 3D reconstruction, and the accuracy of the result will directly affect the accuracy of subsequent quantitative calculations.
    
    For 2D boundary detection, the region of interest (ROI) in poly-crystalline images is the grain boundary. Unfortunately, when etching and polishing samples, inevitable defects such as blurred or missing boundaries and scratches appear in the image and seriously influence the automatical boundary detection, as shown in Figure \ref{microscopic_images}(c) and (d). To tackle this problem, several works have been proposed\cite{feng2017reconstruction,waggoner2015topology-preserving,waggoner20133d,Zhou2014Edge}. Ma \emph{et al.}\cite{ma2019fast-finecut} presented a graph-cut based method, which used the boundary in previous slice ($i$th slice) to recover the boundary in target slice ($(i+1)$th slice), but it limit the topological change of the target image. Convolution neural network (CNN) has driven a great success on image segmentation~\cite{Ni_2019_CVPR} and boundary detection~\cite{he2019bi} in recent years. However, as far as we know, there is no deep learning-based method specially designed for poly-crystalline structural materials with such kind of defects. Another issue existing in boundary detection task is class imbalance, where the amount of grain pixel is much more than that of boundary pixel. The mainstream method is to weight some pixels to promote network balanced learning. Most of them tend to strictly control the position of the boundary in image. However, in materials science, the topology of microscopic images is the most important, and the manual labelled boundaries will inevitably have some errors. Therefore, the algorithm needs to focus on boundary topology while tolerate some offset on boundaries.
    
    In this paper, we propose a novel Weighted Propagation Convolution Neural Network, called WPU-Net, for 3D microscopic image segmentation. To implement slice-wise grain segmentation, the spatial consistency information is utilized and propagated between slices to eliminate the defects in raw images. Besides, we customize adaptive boundary weight for each pixel in each grain to increase the weight of ROI. So that the grain's geometric and topological characteristics can be preserved. Experiments demonstrate that the proposed method achieves promising performance in both of objective and subjective assessment. In 2D boundary detection task, it reduces the error rate by 7$\%$, which outperforms state-of-the-art methods by a large margin. 
    
    In general, our work presents three contributions:
    \begin{itemize}
    \item We introduce spatial consistency information into network architecture, which can eliminate the defects in raw image.
    \item We propose an adaptive boundary weighted loss to preserve geometric and topological characteristics of grain.
    \item We provide a microscopic images dataset with the goal of advancing the state-of-the-art in image processing for materials sciences.\footnote{The code and data are available at \url{https://github.com/clovermini/WPU-Net}.}
    \end{itemize}

	\section{Related Work}
	\label{sec:realted_work}
	\subsection{Boundary Detection}
    \label{sec:boundary detection}
    Most existing boundary detection methods can be categorized into two classes: 2D image-based and 3D image-based methods. 
    
    2D image-based methods detect the boundary only from the 2D image itself, where 2D FCN~\cite{8451775,ma2018deep,Tao2016RDN,zhang2019net,ma2021sesf} has become the de facto standard. U-Net~\cite{ronneberger2015u-net} is one commonly used method for microscopic image segmentation. Many improved methods~\cite{abraham2019novel, Alom2018R2UNet, Oktay2018Attention} are build upon it; 3D image-based methods detect the boundary using the 3D context information contained in the image volume. 3D FCN~\cite{funke2018large,zhu2019v} and RNN are often used for the extraction and fusion of 3D context information, represented by 3D U-Net~\cite{cicek20163d} and UNet+BDCLSTM~\cite{chen2016combining}. In order to solve the problem of the anisotropy in volume image, some researchers combine the 3D convolution with 2D convolution. They connect the 2D convolution layer behind the 3D convolutional layer~\cite{zeng2017deepem3d}, or assign 2D convolution pre-training parameters to 3D convolution by migration~\cite{liu20183d}, etc. Tracking-based strategy is also a common way of using 3D context information. 
    Traditional tracking-based method is used for video object segmentation~\cite{jampani2017video,perazzi2017learning} to maintain temporal consistency. Some works tried to use it to maintain the spatial consistency of material in microscopic image processing. Minnan \emph{et al.}\cite{feng2017reconstruction} proposed an interactive segmentation method based on break-point detection, but a lot of artificial correction is needed. Wangger \emph{et al.}\cite{ma2019fast-finecut,waggoner2015topology-preserving, waggoner20133d,Zhou2014Edge} proposed the concept {\em propagation segmentation} based on graph-cut, which models the spatial relationship between adjacent microscopic images, but it limits the topological change of the target image because hand-crafted features. By virtue of the spatial consistency, it can be possible to use information from one slice to recover the blurred or missing boundary and remove the scratches in adjacent slice. Thus, our method address this task with a tracking-based network.
    
    \subsection{Weighted Loss}
    \label{sec:weighted loss}
    Weighted loss is widely used to handle the class imbalance problem in deep learning. For example, class balanced weighted (CBW) loss~\cite{xie2015holistically-nested} defines the weighted map according to the number of different classes. Lin \emph{et al.}\cite{lin2017focal} has proposed the Focal Loss, which increase the weight of hard samples according to the network result. Both of these methods rigidly force the network to learn the specific location of the object. However, for the boundary detection task in microscopic images, the main concern is the geometric and topological characteristics of grain. CBW and Focal Loss do not consider such information in loss design. U-Net~\cite{ronneberger2015u-net} has proposed a weighted map loss to force the network to learn the separated borders between two regions, which is very suited to loosely arranged regions. However, for tightly arranged regions, $d_1$ and $d_2$ in U-Net are equals to 0 and the result is same with the simplest class balanced weight. Therefore, it is need to design a new weighted loss for tightly arranged regions, such as poly-crystalline structure. %\textcolor[rgb]{1,0,0}{
    The weight maps between different arranged structures can be found at supplemental materials.
    %}

	\section{Method}
	\label{sec:method}
	\subsection{Overview of the Proposed Method}
    \label{sec:overview of the Proposed method}
    \begin{figure*}[htb]
    \centering
    \includegraphics[width=\linewidth]{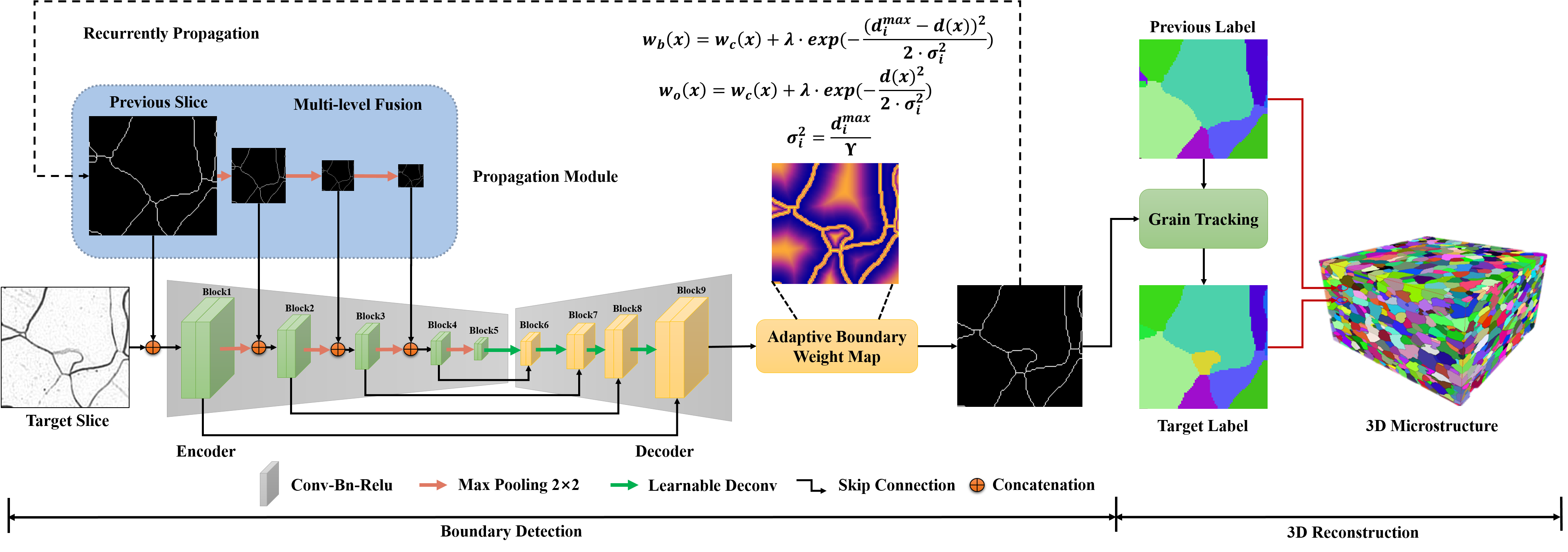}
    \caption{Flowchart of the proposed method.}
    \label{flowchart}
    \end{figure*}
    Figure \ref{flowchart} illustrates the overall flowchart of proposed method. The boundary detection frame is primarily based on an encoder-decoder network, with the propagation module and adaptive boundary weighted loss. The U-Net~\cite{ronneberger2015u-net} is used as the backbone. The propagation module takes the detection result of the previous slice as input and connects it with the feature maps of the target slice to integrate spatial consistency in the network. The adaptive boundary weighted loss customize weight for each pixel in target slice to control the its importance in network learning. As a downstream task, the 3D reconstruction module uses the results of image segmentation for grain track, and restores a series of 2D segmentation results into a 3D voxel form.
    
    \subsection{Integrate Spatial Consistency in Network}
    \label{sec:integrate spatial consistency in network}
    In order to overcome the problem of blurred or missing boundaries in the microscopic images of polycrystalline materials, a propagation segmentation network architecture is proposed. This architecture propagates the boundary detection result from the previous slice to the next target slice for assisting the boundary detection. More specifically, as shown in Figure \ref{flowchart}, the boundary detection result of previous slice is sent to the encoder along with target slice as input. Due to the spatial consistency, the boundary of previous slice has practical guiding significance for the detection of the target slice boundary. If the consistency is strong enough, the boundary of previous slice can even be directly used to recover the blurred or missing boundaries in target slice. The core idea of our work is to build a deep learning model that can use the power of the neural network to learn spatial consistence between two adjacent slices. And the ideal state of our model is that it can not only recognize blurred or missing boundaries and scratches in target slice with the help of the previous slice but also keep the topology of the target slice itself.
    
    To integrate more boundary information from previous slice, a multi-level fusion strategy is presented, as shown in the propagation module in Figure \ref{flowchart}. It means that the boundary result of previous slice is simultaneously connected to the multi-scale feature map on different layers of the encoder, and max-pooling is used to keep the image size consistent. As CNN is a cascaded framework, with the number of convolution layers increases, it extracts high-dimensional information representations gradually. By using the multi-level fusion strategy, network can not only fuse the simple boundary information in previous slice in the head layer, but also better utilize the high-dimensional topology information in the deeper layer, which is important in boundary detection on the polycrystalline image.
    
    \subsection{Adaptive Boundary Weighted Loss}
    \label{sec:adaptive boundary weighted loss}
    Traditional weighted cross entropy rigidly controls the location of the predicted boundary at pixel level, like class balanced weighted (CBW) loss~\cite{xie2015holistically-nested}. The equation is defined as follow:
    \begin{equation}
    \label{eq1}
    E_{cbw} = - \sum_{x \in \Omega} w_c(x) \cdot log(p_{\iota(x)}(x))
    \end{equation}
    Here, $E_{cbw}$ is a pixel-wise cross entropy loss based on softmax. $w_c(x)$ is the weighted value to balance the class frequencies for each pixel $x$.
    $p_{\iota(x)}(x)$ is the activation output from softmax. And $\iota(x) : \Omega \rightarrow \{1,...,K\}$ is the true label of each pixel $x$. $K$ is the number of classes and equals 2 in boundary detection task. Depend on this loss, the CBW can give the certain weighted value to some pixels in training. However, it would lead the network to focusing on the specific location of boundary, while  ignoring the topology of grain.
    
    Therefore, we propose an adaptive boundary weighted (ABW) loss as below:
    % \begin{equation}
    % \begin{aligned}
    % \label{eq2}
    % & E_{abw} = - \sum_{x \in \Omega} \\
    % & \begin{cases} w_{b}(x) \cdot log(p_0(x)), & \mbox{if }w_{b}(x) \geq w_{o}(x) \cdot m_{d}(x) \\ w_{o}(x) \cdot log(p_1(x)), & \mbox{if }w_{b}(x) < w_{o}(x) \cdot m_{d}(x) \end{cases}
    % \end{aligned}
    % \end{equation}
    \begin{equation}
    \begin{aligned}
    \label{eq2}
    & E_{abw} = - \sum_{x \in \Omega}
    & \begin{cases} w_{b}(x) \cdot log(p_0(x))
    \\ w_{o}(x) \cdot log(p_1(x))
    \end{cases}
    \end{aligned}
    \end{equation}
    
    $w_{b}(x)$ and $w_{o}(x)$, are designed for background (label 0) and object (label 1) respectively. $p_0(x)$ and $p_1(x)$ are activation outputs from softmax. %$m_{d}$ is the dilating result of the single-pixel width ground truth boundary which used to control the range of $w_{b} (x)$ and $w_{o} (x)$. %
    The specific definition of $w_{b} (x)$ and $w_{o} (x)$ are shown below:
    
    \begin{equation}
    \label{eq3}
    w_{b}(x) = w_c(x) + \lambda \cdot exp(-\frac{(d_i^{max} - d(x))^2}{2 \cdot \sigma_i^2})
    \end{equation}
    \begin{equation}
    \label{eq4}
    w_{o}(x) = w_c(x) + \lambda \cdot exp(-\frac{d(x)^2}{2 \cdot \sigma_i^2})
    \end{equation}
    \begin{equation}
    \label{eq5}
    \sigma_i = \frac{d_i^{max}}{\gamma}
    \end{equation}
    
    Here, $w_c(x)$ is class balanced weight which is used in CBW. $\lambda$ is the parameter to control the influence of class balanced weight and adaptive boundary weight. For each pixel $x$, $d(x)$ denotes its distance to the nearest boundary and $d_i^{max}$ denotes the maximum of $d(x)$ for all pixels in grain $i$. The standard deviation $\sigma_i$ of normal function in each grain $i$ is the result of $d_i^{max}$ divided by $\gamma$, ($\gamma = 2.58$).
    %\textcolor[rgb]{1,0,0}{
    Thus, by making such customization, the algorithm adaptively controls the weight’s convergence speed with respect to normal distributions. The smaller the grain size, the faster it is converged to $w_c(x)$ at the center of grain on foreground weight map, the faster it is converged to $w_c(x)$ at the boundary of grain on background weight map, which protects the tiny grain and tolerate minor differences in boundary location.
    %}
    Then, for each pixel $x$, the ABW find its corresponding grain $i$, $d_i^{max}$ and $\sigma_i$, and then calculate its weights ($w_{b} (x)$ and $w_{o} (x)$). 
    
    \begin{figure}[htb]
    \centering
    \includegraphics[width=1.0\linewidth]{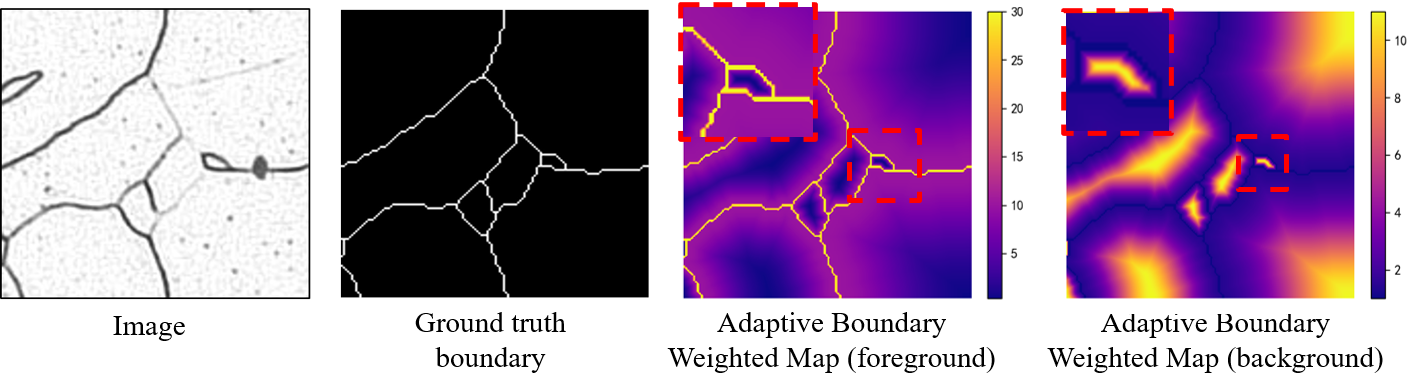}
    \caption{Demonstration of the adaptive boundary weighted map. From left to right are raw image, boundary result, object weight map and background weight map. A tiny grain is specifically marked with red rectangle and magnified to visually display how adaptive weighting protects tiny grains. The smaller the grain size, the faster the weight converge.}
    \label{fig2}
    \end{figure}
    
    The basic principle of these weights is to model the topology of grain boundary. That is the region is likely to be boundary if it's close to ground truth boundary, and $w_{o} (x)$ inclines to be larger. While the region is likely to be grain if it's close to the center of grain, and $w_{b} (x)$ inclines to be larger. The adaptive boundary weighted map is shown in Figure \ref{fig2}. And one tiny grain is magnified with red rectangle in order to show the effect of topology preservation.

	\section{The pure iron grain dataset}
    \label{sec: the pure iron grain dataset}
    
    Recently progress in image segmentation has been driven by high-capacity models trained on large dataset. However, the production and labelling of material microscopic image are very time-consuming, and there are rarely published material dataset currently. The lack of large-scale image data has seriously hampered the development of materials science. Therefore, we open our labelled dataset in order to provide a referenced dataset for computer vision community. 
    
    The dataset which we provided is produced and collected in practical experiments with serial section method, which consists of raw images (shown in Figure \ref{microscopic_images}(c) and (d)), and their ground truth boundary and label results. The ground truth boundaries and 3D grain labels for dataset are labelled by professional material researchers. It is a stack of 296 microscopic pure iron slices with large size ($1024\times 1024$ pixels) for two classes (grain and grain boundary). Typically, the image processing process includes detecting grain boundaries in each slice and reconstructing the 3D microstructure. It's worth noting that, due to the polishing process of sample preparation, the resolution of Z direction is always smaller than X and Y direction, as shown in Figure \ref{microscopic_images}(b).
    	
	\section{Experiment Results}
    \label{sec: experimental results}
    
    In this section, adequate experiments will be deployed to demonstrate the effectiveness of our proposed method. The real dataset is divided into a training set, a validation set, and a test set, each containing consecutive 116, 32, and 148 images.  To ensure sufficient training data, we perform random cropping, flipping and rotation of the data during the training process. And random seeds are set for the repeatability of experiments. For the test set and validation set, images are cut by overlap-tile strategy~\cite{ma2018deep, ma2020data, ma2022end}, from $1024 \times 1024$ size to $256 \times 256$ size. 
    
    The goal of boundary detection in this work is to obtain single-pixel closed boundary of each grain in the image, where the grain topology is most important. Therefore, the metric should tolerate minor differences in boundary location between prediction and ground truth and pay attention to under-segment and over-segment errors. For fair comparison, three metrics are used for evaluation, variation of information (VI)~\cite{meil2007comparing,nuneziglesias2013machine}, adjusted rand index (ARI)~\cite{vinh2010information} and mean average precision (mAP)~\cite{lin2014microsoft,kaggle2018dsb}. For VI metric, a lower value indicates a better performance. While for the other metrics, a larger value indicates a better performance.
    
    Input images are normalized first. The weights of nets are initialized with Xavier~\cite{glorot2010understanding} and all nets are trained from scratch. We adapt batch normalization (BN)~\cite{ioffe2015batch} after each convolution and before activation. All hidden layers are equipped with Rectified Linear Unit (ReLU~\cite{krizhevsky2012imagenet}). The learning rate is set to 1e-4 initially, decaying by 80 percent per 10 epochs until 1e-6. We optimize the objective function with respect to the weights at all network layer by RMSProp with smoothing constant $\alpha$=0.9 and $\epsilon$=1e-5. Each model is trained for 500 epochs on 1 Tesla V100 GPU (32GB) on Nvidia DGX Station with a batch size of 24. $\lambda$ is set to 10 in Eq. \ref{eq3} and \ref{eq4}.  
    %$5 \times 5$ circle structural kernel is applied for dilation in $m_{d}$ of Eq. \ref{eq2}. %
    In order to obtain a single-pixel boundary result, all networks predictions undergo a skeletonization operation. All parameters above are selected by grid search. We implement our method using PyTorch~\cite{paszke2019pytorch}. During training, we pick up the best parameter when network achieves the smallest loss on the validation set. All the performance in experiment is obtained on the test set using the best parameter. 
    
    % \textbf{0.3028}  \multicolumn{1}{l|}{0.3114}
    \begin{table*}[htb]
    \begin{center}
    \begin{tabular}{|c|c|c|c|c|c|c|c|}
    % {\hsize}{@{}@{\extracolsep{\fill}}cccccccc@{}}
    %
    \hline
    \multirow{2}*{Algorithm}&\multicolumn{2}{c|}{Loss}&\multicolumn{2}{c|}{Mode}&\multirow{2}*{VI $\downarrow$ }&\multirow{2}*{MAP $\uparrow$}&\multirow{2}*{ARI $\uparrow$} \\
    \cline{2-5}
    &CBW&ABW&F1&F1-4&&& \\\hline
    \multirow{2}{*}{U-Net}& $\checkmark$&$\times$&$\times$&$\times$& 0.3165 & 0.6067 & 0.8155 \\ 
    \cline{2-8} 
    & $\times$ & $\checkmark$ & $\times$ & $\times$ & 0.2952 & 0.6171 & 0.8297 \\\hline 
    \multirow{2}{*}{\begin{tabular}[c]{@{}c@{}}Attention\\ U-Net\end{tabular}}  & $\checkmark$  & $\times$ & $\times$ & $\times$  & 0.2828    & 0.6264 & 0.8298   \\
    \cline{2-8} 
    & $\times$ & $\checkmark$ & $\times$ & $\times$ & 0.2808 & 0.6298   & 0.8370 \\\hline 
    \multirow{4}{*}{WPU-Net}  & $\checkmark$  & $\times$ & $\checkmark$ & $\times$ & \textbf{0.1868} & 0.6703 & 0.8530\\
    \cline{2-8} 
    & $\checkmark$ & $\times$ & $\times$ & $\checkmark$ & 0.1894 & 0.6718 & 0.8516 \\
    \cline{2-8} 
    & $\times$ & $\checkmark$ & $\checkmark$ & $\times$ & 0.1929 & 0.6925 & 0.8634 \\
    \cline{2-8} 
    & $\times$ & $\checkmark$ & $\times$ & $\checkmark$ & 0.1874 & \textbf{0.6959} & \textbf{0.8647} \\\hline
    \end{tabular}
    % }
    \end{center}
    \caption{%\textcolor[rgb]{1,0,0}{
    Ablation experiments results on loss and mode. WPU-Net is composed of U-Net and propagation module. The bold values mean the best performance in each metric.
    %}
    }
    \label{table1}
    \end{table*}
    
    All reported performance is the average of scores for all images in test set. To justify the effectiveness of our proposed method, we have conducted a sufficient ablation experiment, the results are shown in Tables \ref{table1} and \ref{table3}. The following is detailed analysis and explanation of the experimental results.
    
    \subsection{Adaptive Boundary Weighting}
    \label{sec:adaptive boundary weighting experiment}
    
    As shown in Table \ref{table1}, adaptive boundary weight (ABW) performs better than class-balanced weight (CBW) in general, especially on mAP and ARI. We can see the VI score of CBW and ABW may behave differently on WPU-Net with F1 mode, however, the mAP and ARI score of ABW always perform well. This may benefit from the excellent performance of ABW on small grains. As VI is less sensitive in tiny grains. We can also see that ABW can achieve higher performance both on U-Net and Attention U-Net architecture. That is suggesting that improvements induced by ABW can be used with existing state-of-the-art architectures.
    
    \subsubsection{Integrate Spatial Consistency in Network}
    \label{sec:integrate spatial consistency experiment}
    
    %\textcolor[rgb]{1,0,0}{
    To systematically examine the effect of the proposed propagation module, we conduct the ablation experiment on WPU-net. WPU-Net is composed of U-Net and propagation module. Owing to the integration of spatial consistency by the proposed propagation module, WPU-net achieves a promising improvement in performance compared to the U-net. In addition, we further analyze two different fusion modes for spatial information, F1 and F1-4. F1 means the previous slice’s information is only merged in the head layer. While F1-4 means the multi-scale fusion strategy. We can see from Table \ref{table1} that the results with multi-level fusion are generally better. It proves that multi-scale fusion could make better use of rich information between layers.
    %}
    
\begin{figure}[htb]
    \centering
    \includegraphics[width=0.8\linewidth]{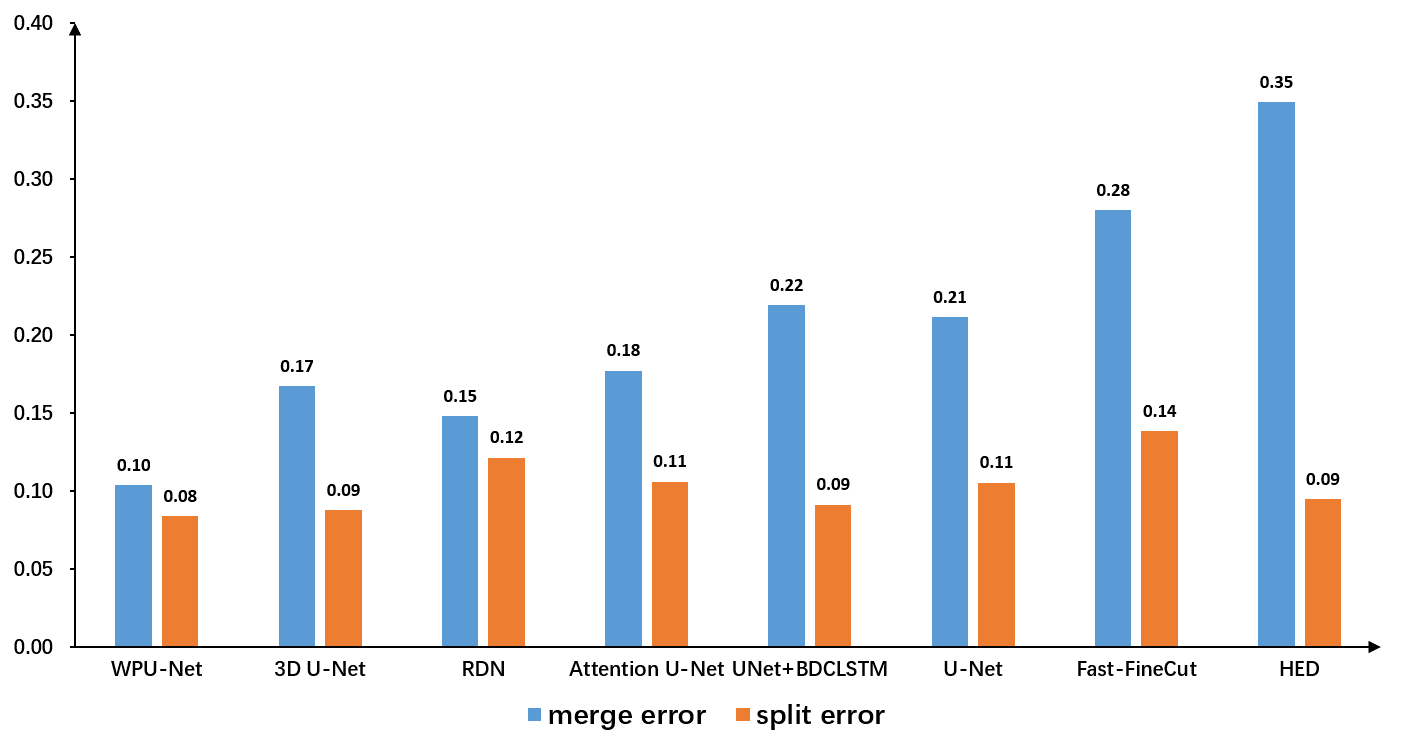}
    \caption{Model comparison on merge error and split error.}
    \label{fig7}
    \end{figure}

\begin{figure*}[t]
    \centering
    \includegraphics[width=1.0\linewidth]{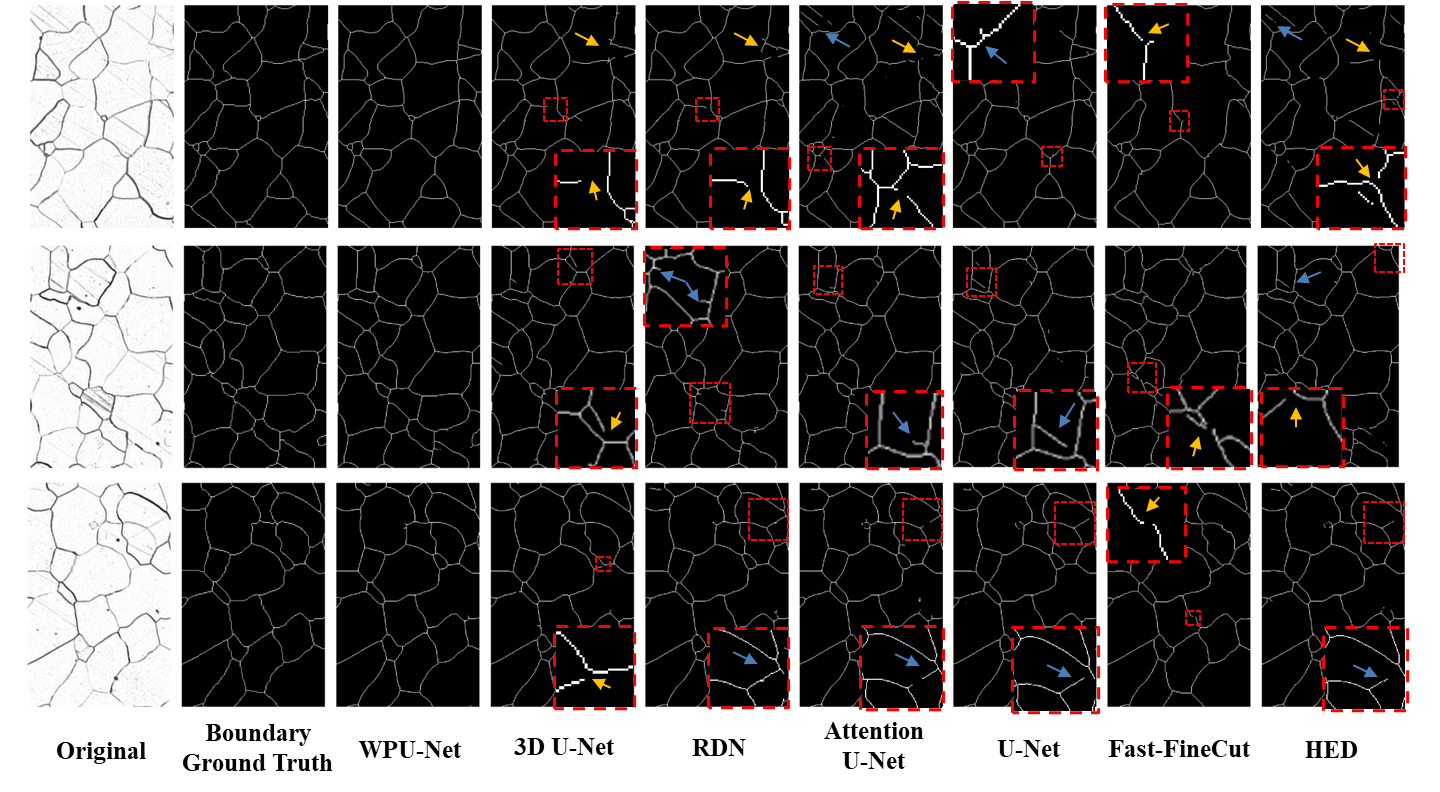}
    \caption{Boundary Detection results with different methods. Three slices from top to bottom. The yellow arrows represent blurred or missing boundary that does not recover and detect accurately. The blue arrows refer to scratches that does not eliminate. And some defective regions are magnified and shown in red rectangles. Our proposed method performs better compared to other state-of-the art methods.}
    \label{fig9}
    \end{figure*}

    Another experiment we conducted is a model comparison between WPU-Net and seven state-of-the-art methods, which are 3D U-Net~\cite{cicek20163d}, Attention U-Net~\cite{Oktay2018Attention}, RDN~\cite{Tao2016RDN}, U-Net~\cite{ronneberger2015u-net}, UNet+BDCLSTM~\cite{chen2016combining}, HED~\cite{xie2015holistically-nested} and Fast-FineCut~\cite{ma2019fast-finecut}. Table \ref{table3} and Figure \ref{fig7} show the quantitative assessment of above methods. And the visualization of boundary detection results is shown in Figure \ref{fig9}.
    
    \begin{table}[htb]
    \begin{center}
    \begin{tabular}{|c|c|c|c|c|}
    \hline
    Algorithm       & VI $\downarrow$ & MAP $\uparrow$  & ARI $\uparrow$   \\ \hline
    WPU-Net         & \textbf{0.1874} & \textbf{0.6959} & \textbf{0.8647}   \\ \hline
    3D U-Net        & 0.2537          & 0.6496          & 0.8397            \\ \hline
    RDN             & 0.2691          & 0.6340          & 0.8418            \\ \hline
    Attention U-Net & 0.2828          & 0.6264          & 0.8298            \\ \hline
    UNet+BDCLSTM    & 0.3134          & 0.6193          & 0.8135            \\ \hline
    U-Net           & 0.3165          & 0.6067          & 0.8155            \\ \hline
    Fast-FineCut    & 0.4183          & 0.5660          & 0.8030            \\ \hline
    HED             & 0.4436          & 0.5651          & 0.7635            \\ \hline
    \end{tabular}
    \end{center}
    \caption{Model comparison for boundary detection. The bold values mean the best performance in each metric.}
    \label{table3}
    \end{table}
    
    \begin{figure*}[t]
    \centering
    \includegraphics[width=1.0\linewidth]{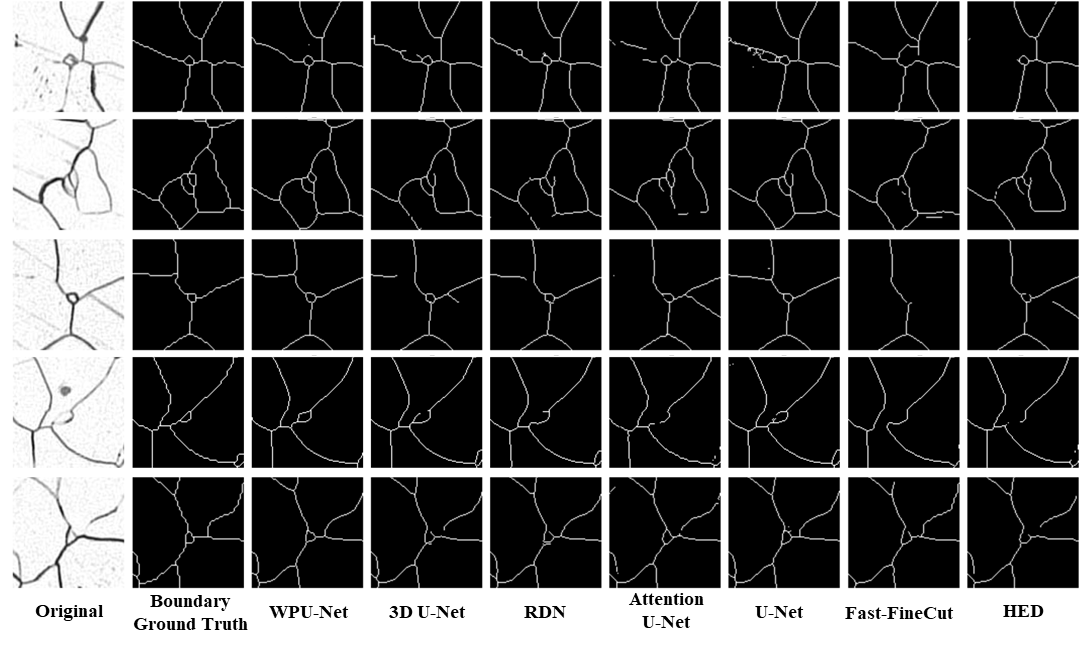}
    \caption{Qualitative results of tiny grains with different segmentation methods.}
    \label{fig10}
    \end{figure*}
    
    As shown in Table \ref{table3}, 3D U-net achieves the second-best performance in our experiment which can utilize 3D information in the volume. However, the 3D convolutional kernel and LSTM have numerous parameters, therefore require much more data to train and is easily inclined to over-fitting. Our proposed method WPU-Net uses propagation strategy to integrate 3D information and only add small parameters. According to the experiment, the WPU-net outperforms others in every evaluation metrics, especially on VI metrics (the summation of split error and merge error), our method is about $7\%$ lower than other methods. This proves the feasibility and effectiveness of WPU-Net in the boundary detection task of 3D polycrystalline material image segmentation. The problem of blurred or missing boundaries and scratches in image are the two main reasons for the in-applicability of typical methods. 
    
    To further analyze it, the merge error and split error of each method in VI evaluation metric are displayed separately in Figure \ref{fig7}. The merge error(under-segmentation) means the error caused by unsuccessful detection of grain boundaries(FN), resulting in two grains in the image being judged to be one grain (usually occurs at blurred boundaries). While the split error (over-segmentation) means the wrong detection of grain boundaries(FP), resulting in one grain in the image is judged as two grains (usually occurs at scratches). In Figure \ref{fig7}, we can see that other models all show much worse performance on blurred boundaries generally. However WPU-Net performs better in both problems, especially on blurred boundaries, as shown in Figure \ref{fig9}. 
    
    %\textcolor[rgb]{1,0,0}{
    In addition, we demonstrate qualitative results for tiny grain in Figure \ref{fig10}. It is shown that our proposed method can correctly segment small grain compared to traditional methods.
    %}
    \begin{figure}[h]
    \centering
    \includegraphics[width=\linewidth]{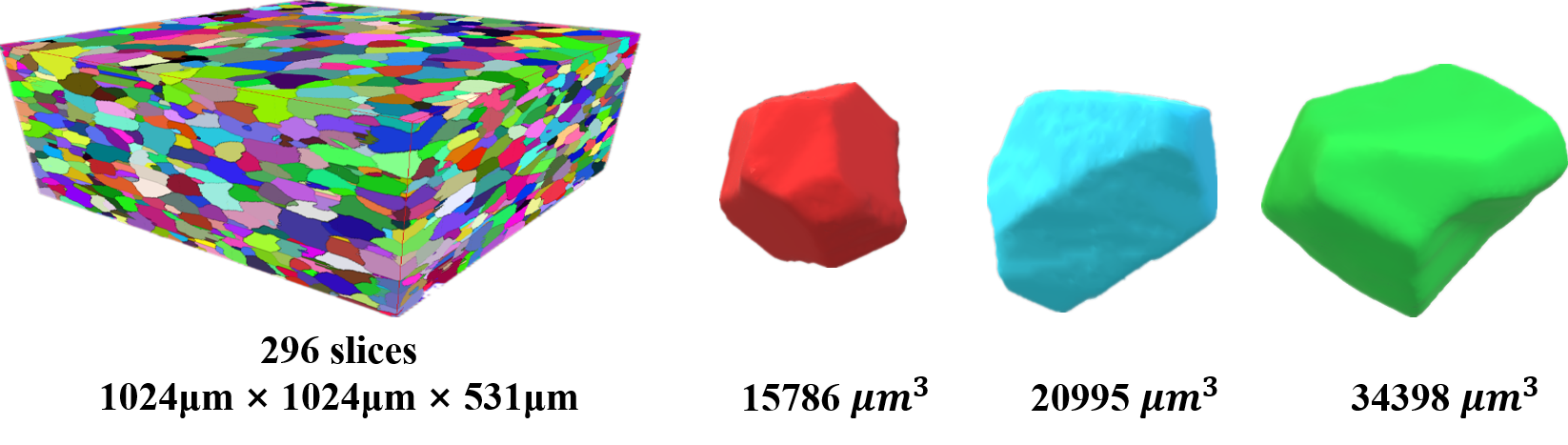}
    \caption{3D visualization of overall and small grains. The upper is 3d structure of real data. The three small grains were re-sized at lower region for better visualization. }
    \label{fig8}
    \end{figure}

    \section{3D reconstruction}
    \label{sec:3d reconstruction}
    After extracting all grain boundaries and tracking all objects, 3D reconstruction is performed on serial slices. The visualization of overall and small grains is shown in Figure \ref{fig8}. In addition, the overall and small grains visualization are supported by Avizo~\cite{avizo8} and meshlab~\cite{meshlab} respectively. The overall 3D reconstruction of real data is $1024 \mu m \times 1024 \mu m \times 531 \mu m$. Besides, four small grains are shown in the lower region of Figure \ref{fig8} with its volume size.
    
    %------------------------------------------------------------------------
    \section{Conclusion}
    \label{sec:conclusion}
    In this work, we propose a Weighted Propagation U-Net (WPU-Net) to address the 3D image segmentation in polycrystalline materials. The network integrates spatial consistency from adjacent slices to aid boundary detection in target slice. And we present adaptive boundary weighting to optimize the model, which can tolerate minor difference in boundary detection and protect the topology of grains. In addition, we provide a material microscopic images dataset with the goal of advancing the state-of-the-art in image processing for materials science. Experiments have shown that our network achieves better performance than previous state-of-the-art methods. In boundary detection task, the VI metric of our method is about 7\% lower than the second-best method. The accurate boundary detection result improves the stability of the 3D reconstruction result as a whole.

	\section{Acknowledgment}
    This work was supported by the National Key Research and Development Program of China under Grant 2020YFB0704501, the National Natural Science Foundation of China under Grant 62106019, China Postdoctral Science Foundation under Grant 2021M700383, Scientific and Technological Innovation Foundation of Shunde Graduate School of USTB under Grant BK20AF001 and Grant BK21BF002, and Fundamental Research Funds for the Central Universities of China under Grant 00007467, and Postdoctor Research Foundation of Shunde Graduate School of University of Science and Technology Beijing under Grant 2021BH005. The computing work is supported by USTB MatCom of Beijing Advanced Innovation Center for Materials Genome Engineering. In addition, we thank LetPub for its linguistic assistance during the preparation of this manuscript.
	
	\bibliographystyle{elsarticle-num}
    \bibliography{egbib.bib}

\end{document}